\documentclass[times, preprint, 10pt]{elsarticle}
\usepackage{algorithm}
\usepackage{algpseudocode}
\usepackage{amsmath,amssymb}
\usepackage{graphicx}
\graphicspath{{figs/}{./}}
\usepackage{tikz}
\usepackage{booktabs}
\usepackage{multirow}
\usepackage{url}
\usepackage{hyperref}
\usepackage{textcomp}
\usepackage{bbm}
\usepackage{lineno}
\usepackage{setspace}
\usepackage{xcolor}
\usetikzlibrary{positioning,arrows.meta,shapes.geometric,fit,backgrounds,calc}
\usepackage{xspace}

\usepackage{etoolbox}

\biboptions{sort&compress}

\AtBeginEnvironment{figure}{\singlespacing}
\AtBeginEnvironment{figure*}{\singlespacing}
\AtBeginEnvironment{table}{\singlespacing}
\AtBeginEnvironment{table*}{\singlespacing}
\AtBeginEnvironment{algorithm}{\singlespacing}
\AtBeginEnvironment{thebibliography}{\singlespacing\small}
\begin{document}
\begin{frontmatter}
\title{SSM-Attack: Learning Where to Perturb for Black-Box Face Privacy}
\author[a]{Tunahan Parlay{\i}c{\i}\corref{cor1}\fnref{fn1}}
\ead{tunahanp@sabanciuniv.edu}
\author[a]{Erchan Aptoula}
\author[b]{Y\"{u}cel Sayg{\i}n}
\cortext[cor1]{Corresponding author}
\fntext[fn1]{ORCID: 0000-0001-7250-6375}
\address[a]{Computer Vision and Pattern Analysis Laboratory (VPALab), Faculty of Engineering and Natural Sciences, Sabanci University, Istanbul, T\"{u}rkiye}
\address[b]{Faculty of Engineering and Natural Sciences, Sabanci University, Istanbul, T\"{u}rkiye}
\begin{abstract}
Face recognition systems are deployed widely enough that a photograph shared
online can be matched without its subject's consent. Generator-based protection
methods rely on heavy pretrained backbones, multi-stage inference or visible
semantic edits. We introduce SSM-Attack, a black-box face privacy framework built
on Selective State Space Models. Rather than synthesizing the perturbation, the
state-space model predicts in one linear-time pass a \emph{spatial allocation} of
the perturbation budget, marking which facial locations carry identity evidence;
a short projected refinement then sets the values within this per-pixel budget.
Against published benchmark results, \textbf{SSM-Attack attains the highest
average black-box attack success rate on all three evaluation sets: 87.10
percent on CelebA-HQ, 86.90 on LADN and 81.15 on FFHQ}, the last from a network
that never saw that dataset. The allocation network holds 5.73 million
parameters, or 16.13 million with the frozen networks used by its quality terms.
Under the same cap of 0.10 on the maximum per-pixel change,
allocation improves structural similarity by 0.044, peak signal-to-noise ratio
by 1.62 decibels and Fr\'{e}chet Inception Distance by 11.2 points over a
uniform map while spending 37 percent of its capacity. Strong facial privacy
protection therefore does not require a large generative prior; it requires
knowing where to spend a small one.
\end{abstract}
\end{frontmatter}
\section{Introduction}
\label{sec:intro}
Face recognition (FR) systems, powered by discriminative margin-based losses~\citep{deng2019arcface,schroff2015facenet} and increasingly deep architectures, are now deployed at scale in surveillance, biometric authentication, access control and social media tagging. Facial images are captured, stored and analyzed in these settings without explicit consent, so individuals can be tracked across public spaces, profiled from their photographs, or exposed to identity fraud~\citep{cherepanova2021lowkey, shan2020fawkes}. Adversarial perturbations offer a proactive, model-agnostic defense: a signal added to an image before it is shared disrupts FR inference while the picture still looks like the person~\citep{goodfellow2015explaining, szegedy2014intriguing}.
Early white-box attacks such as FGSM~\citep{goodfellow2015explaining} and PGD~\citep{madry2018pgd}, and the transferability extensions built on them~\citep{dong2018boosting,dong2019evading,xie2019difgsm}, remain limited by pixel-local gradient optimization, which overfits to surrogate-specific texture cues and transfers poorly to architecturally distinct targets.
To overcome this transferability bottleneck, recent research has shifted toward generator-based methods that synthesize identity-disruptive perturbations through GAN-based makeup transfer~\citep{huang2022amtgan,shamshad2023clip2protect}, diffusion-based generation~\citep{sun2024diffam}, and portrait pipelines~\citep{wang2025advcpg}. Despite achieving high attack success rates, these \emph{unrestricted} methods share common drawbacks: they require heavy pretrained backbones ($>$100M--1B+ parameters), multi-stage inference, and produce visible semantic edits. This raises a natural question: \emph{can we achieve comparable transferability within a strict noise budget using an efficient generator?}

Here, we introduce \textbf{SSM-Attack}, a noise-based adversarial face privacy framework built upon Selective SSMs~\citep{gu2021s4,gu2023mamba}. Mamba~\citep{gu2023mamba} models long-range dependencies through input-dependent selective state updates at a cost linear in the number of spatial tokens, without the local receptive field of a convolution or the quadratic cost of attention. Our central design choice concerns \emph{what} the state-space model is asked to produce. Rather than emitting the perturbation itself, the network processes the face as a sequence of 256 spatial tokens and emits a \emph{spatial budget map}: a mask indicating which locations carry identity evidence. The per-pixel $\ell_\infty$ budget is allocated in proportion to this map under a hard global cap, and the perturbation values are then determined by a short projected optimization. The state-space model thus contributes what a sequence model is suited for, carrying context along the token sequence to decide \emph{where} the budget should be spent, while the perturbation itself remains free of generator-induced structure.

We evaluate under a strict Leave-One-Out (LOO) black-box protocol on CelebA-HQ~\citep{karras2018celeba} and LADN~\citep{gu2019ladn}, with cross-dataset validation on FFHQ~\citep{karras2019style}. Every reported number comes from an allocation network whose training data is disjoint from the evaluation set in question (Section~\ref{subsec:exp_setup}), so all results measure generalization to unseen images and unseen identities. SSM-Attack attains the highest average attack success rate (ASR) on all three sets with $7\times$ fewer protection-time parameters than DiffAM and no multi-step diffusion inference.

One methodological distinction matters here. SSM-Attack optimizes a purely \emph{untargeted} dodging objective, whereas most competing methods redirect identity toward a pre-selected target. For privacy the goal is to prevent recognition, not to impersonate, and the two are not equivalent: Zhou et~al.~\citep{zhou2024rethinking} show that in the black-box setting a successful impersonation does not guarantee a successful dodge, since an example pushed toward a target identity can still match its source. We optimize Eq.~\eqref{eq:lid} directly at a false acceptance rate (FAR) of $0.01$.

Our main contributions are:
\begin{itemize}
    \item \textbf{Strong black-box transferability from a compact
    model.} Against published benchmark results, SSM-Attack attains the highest average ASR on \emph{all three} evaluation sets, $87.10\%$ on CelebA-HQ,
    $86.90\%$ on LADN and $81.15\%$ on FFHQ, surpassing the previous best on each
    benchmark (Adv-CPG by $+7.82$ points on CelebA-HQ, the best reported baseline by $+6.93$ points on LADN, Adv-CPG by $+1.13$ points on FFHQ) while using a generator more than two orders of magnitude smaller than the largest baselines.
    \item \textbf{A new role for sequence models in adversarial attacks.} We
    introduce the first SSM-based face privacy framework and identify the
    productive role of a state-space model here: not synthesizing perturbations,
    but \emph{allocating the perturbation budget spatially}, a decision that
    rewards the long-range, linear-time reasoning SSMs provide.
    \item \textbf{Learned allocation shifts the attack--fidelity--budget trade-off.}
    Under the same hard $\ell_\infty$ bound it improves the structural similarity
    index (SSIM, $+0.044$), the peak signal-to-noise ratio (PSNR, $+1.62$\,dB)
    and the Fr\'{e}chet Inception Distance (FID, $-11.2$) over a uniform map at
    $37\%$ of its budget,
    costing $1.84$ points of ASR; a fixed elliptical prior at comparable allocated capacity
    recovers less than half the attack performance, so the allocation is learned
    rather than assumed.
\end{itemize}

\section{Related Work}
\label{sec:related}
\noindent\textbf{Noise-based adversarial face privacy.} Early face privacy methods relied on bounded adversarial noise to disrupt the embedding spaces of deep FR models trained with discriminative margin losses~\citep{szegedy2014intriguing,goodfellow2015explaining,deng2019arcface}. While fundamental optimization techniques like PGD~\citep{madry2018pgd}, momentum (MI-FGSM)~\citep{dong2018boosting}, and translation-invariant smoothing (TI-DIM)~\citep{dong2019evading,xie2019difgsm} improved general transferability, identity in FR relies on holistic representations. Consequently, pixel-local attacks often overfit to surrogate-specific texture biases~\citep{zhou2025dpa}. Face-specific methods narrowed this gap: TIP-IM~\citep{yang2021tipim} optimizes identity masks directly, Fawkes~\citep{shan2020fawkes} poisons the enrolment stage, LowKey~\citep{cherepanova2021lowkey} adds a frequency-domain constraint, and reversible formulations~\citep{liu2023unauthorized} let the owner recover the original image.

To overcome the limitations of localized noise, unrestricted methods alter facial semantics using generative priors. Adversarial makeup transfer organizes perturbations into plausible cosmetic patterns to suppress identity~\citep{yin2021advmakeup,huang2022amtgan,shamshad2023clip2protect,li2024gift,shamshad2024dfpp}, while recent work employs diffusion models~\citep{sun2024diffam,liu2024advdiffusion,liu2025diffprotect} or integrates adversarial objectives into large-scale portrait generation~\citep{wang2025advcpg}. Despite high attack success rates, these approaches require multi-stage inference, heavy pretrained backbones, or semantic edits that visibly alter facial appearance~\citep{komkov2021advhat,liu2025advcloak}.

While Vision Transformers (ViT~\citep{dosovitskiy2021vit}) offer the global dependency modeling necessary for effective identity suppression, their quadratic self-attention complexity scales poorly with high-resolution facial images.
Structured state space models~\citep{gu2021s4} offer linear-time sequence modeling instead, and Mamba~\citep{gu2023mamba} makes their state update input-dependent, so that what a position retains is decided from its own content.
Recent adaptations of Mamba for visual tasks~\citep{zhu2024visionmamba} have demonstrated a strong capacity for long-range interactions with improved scalability.
To the best of our knowledge, our work is the \emph{first} to use SSMs for face privacy, and it uses them to decide where a bounded perturbation is spent in strict black-box FR settings rather than to synthesize it.
\section{Methodology}
\label{sec:methodology}
\begin{figure*}[t]
\begin{center}
\makebox[\textwidth][c]{\includegraphics[width=1.25\textwidth]{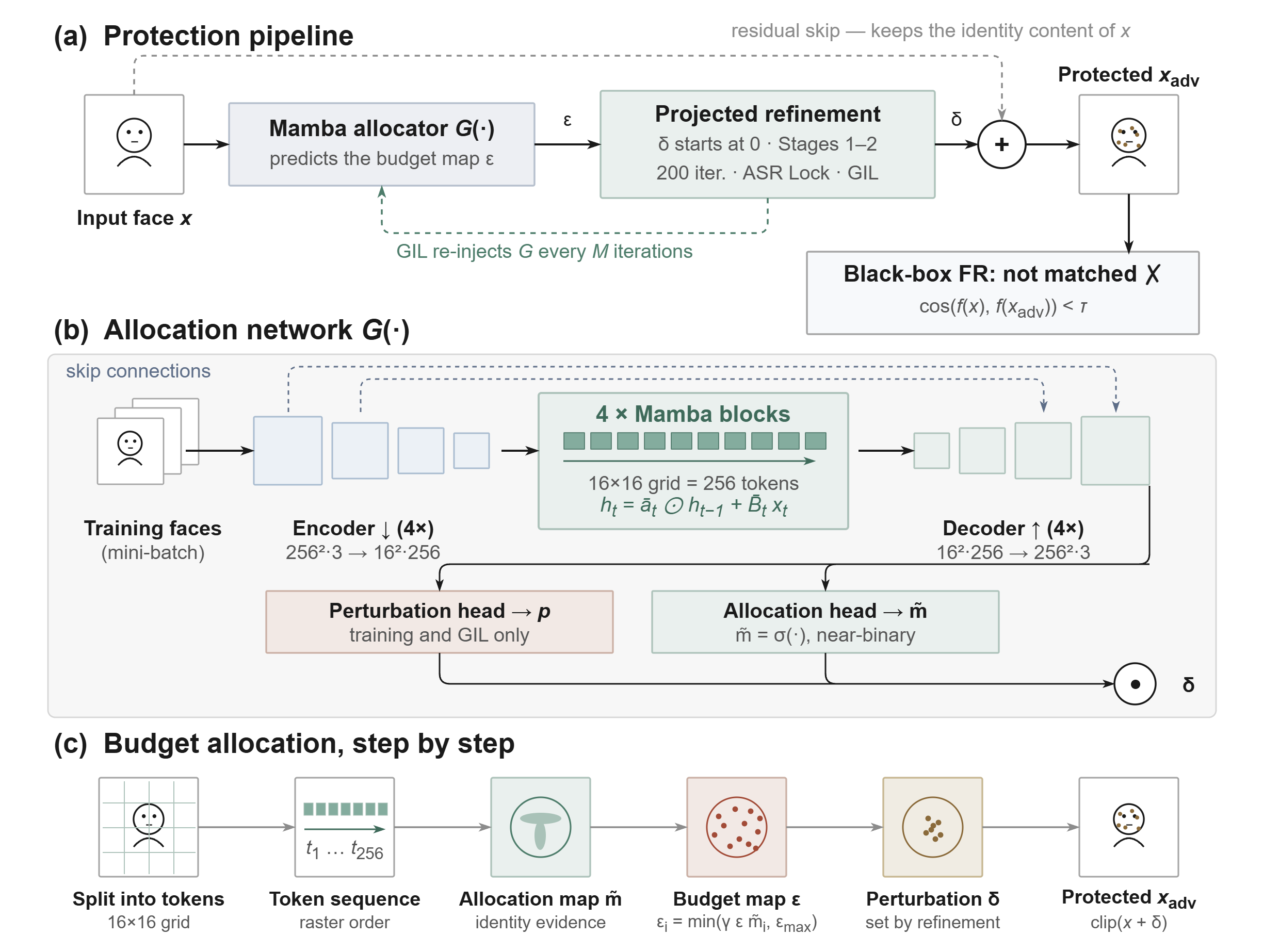}}
\end{center}
\caption{\textbf{Overview of SSM-Attack.} The network predicts \emph{where}
the $\ell_\infty$ budget should be spent; a projected refinement then decides the
values. \textbf{(a)} One forward pass of $G(\cdot)$ gives the budget map
$\boldsymbol{\epsilon}$ (Eq.~\eqref{eq:budget_map}); from
$\boldsymbol{\delta}=\mathbf{0}$, 200 iterations run under the ASR-Lock schedule
with periodic GIL corrections, each projected element-wise onto
$\boldsymbol{\epsilon}$ so that $\|\boldsymbol{\delta}\|_\infty \le
\epsilon_{\max}$ holds by construction. $\tau$ is the verification threshold at
FAR@0.01. \textbf{(b)} A strided encoder maps $\mathbf{x}$ to $N{=}256$ tokens,
four Mamba blocks scan them in linear time, and a decoder with U-Net skips feeds
the allocation head; the perturbation head never sets $\boldsymbol{\delta}$ and
acts only through GIL. \textbf{(c)} The mask is scaled by $\gamma$ and capped at
$\epsilon_{\max}{=}0.10$.}
\label{fig:overview}
\end{figure*}

SSM-Attack operates entirely in a bounded $\ell_\infty$ perturbation space, producing additive pixel-level noise rather than semantic edits.
As illustrated in Fig.~\ref{fig:overview}, it combines a Mamba-based allocation network $G(\cdot)$ with dual output heads, a two-stage curriculum with adaptive ASR Lock, and Generator Injection Learning (GIL) for escaping local minima during projected refinement.

\subsection{Problem Formulation}
\label{subsec:problem}
Given a face image $\mathbf{x} \in \mathbb{R}^{3 \times H \times W}$ normalized to $[0,1]$, we seek an additive perturbation $\boldsymbol{\delta} \in \mathbb{R}^{3 \times H \times W}$ such that the adversarial image
\begin{equation}
    \mathbf{x}_{\text{adv}}
        = \max\!\bigl(\mathbf{0},\, \min(\mathbf{1},\, \mathbf{x} + \boldsymbol{\delta})\bigr)
    \label{eq:xadv}
\end{equation}
evades FR systems in a black-box setting while remaining visually close to the original.
Here, $\boldsymbol{\delta}$ is the pixel-domain perturbation we optimize, and the clamp in Eq.~\eqref{eq:xadv}, which we abbreviate as $\mathrm{clip}(\cdot,0,1)$ below, keeps each pixel in the valid intensity range $[0,1]$.

Let $\{f_1, \ldots, f_K\}$ be a set of FR models, where each $f_k : \mathbb{R}^{3 \times H \times W} \to \mathcal{S}^{d-1}$ maps a face image to an $\ell_2$-normalized identity embedding on the unit hypersphere $\mathcal{S}^{d-1} \subset \mathbb{R}^d$. Each model $f_k$ is associated with a verification threshold $\tau_k$ calibrated at FAR@0.01~\citep{deng2019arcface}.
The attack on a target model $f$ is considered successful for an image $\mathbf{x}$ if $\cos(f(\mathbf{x}),\, f(\mathbf{x}_{\text{adv}})) < \tau_f$. For a held-out target $f$ evaluated over the test set $\mathcal{D}$, the attack success rate~\citep{sharif2019general} is defined as:
\begin{equation}
    \mathrm{ASR}(f)
    = \frac{1}{|\mathcal{D}|}
      \sum_{\mathbf{x}\in\mathcal{D}}
      \mathbbm{1} \Bigl[
          \cos \bigl(f(\mathbf{x}),\, f(\mathbf{x}_{\text{adv}})\bigr)
          < \tau_f
      \Bigr] \in [0, 1],
    \label{eq:asr-final}
\end{equation}
where $\mathbbm{1}[\cdot]$ equals $1$ if its argument is true and $0$ otherwise.

We maximize ASR subject to a bounded pixel-space perturbation. Unlike prior work that enforces a single scalar bound, we enforce a \emph{per-pixel} bound $|\delta_i| \le \epsilon_i$ given by a learned budget map $\boldsymbol{\epsilon}$, itself subject to a hard global cap $\epsilon_{\max} = 0.10$; the conventional scalar constraint is the special case $\epsilon_i \equiv \epsilon_{\max}$. We describe this as a \emph{bounded} rather than an imperceptible perturbation: as Yang et~al.~\citep{yang2021tipim} observe, $\ell_p$-bounded noise at these magnitudes does not align perfectly with human perception, so we report SSIM, PSNR and FID rather than relying on a perceptibility claim. This constraint is enforced throughout by (i)~$\tanh(\cdot)$ scaling at the generator output (Section~\ref{subsec:generator}), (ii)~iterative $\ell_\infty$ projection (denoted $\Pi_{\boldsymbol{\epsilon}}$, which clips each element to the $[-\epsilon, \epsilon]$ range) during refinement, and (iii)~final clipping of $\mathbf{x}_{\text{adv}}$.

To promote black-box transferability, we optimize against an ensemble of $K$ surrogate models (i.e., FR models available during training that serve as proxies for the unknown held-out target) using the identity loss:
\begin{equation}
\mathcal{L}_{\text{id}} = \frac{1}{K}\sum_{k=1}^{K}\cos(f_k(\mathbf{x}),\, f_k(\mathbf{x}_{\text{adv}})),
\label{eq:lid}
\end{equation}
which encourages perturbations that suppress identity cues shared across architectures and training objectives, improving generalization to unseen targets under the LOO protocol~\citep{yang2021tipim,huang2022amtgan}.

\noindent\textbf{Threat model.} The protector is the owner of the image and may modify only that image, once, before it is published. The recognition system that will process it afterwards is unknown: its weights, architecture, training objective and operating threshold are unavailable, and it cannot be queried, so no feedback from the target is used at any stage of training or protection. What the protector does have is a small set of publicly released FR models, used as surrogates. This is the standard setting of the face privacy literature~\citep{yang2021tipim,huang2022amtgan,sun2024diffam,wang2025advcpg} and is stricter than score- or decision-based black-box attacks, which assume query access to the target.
\subsection{Mamba-Based Allocation Network}
\label{subsec:generator}
What the network has to produce dictates its shape. An allocation decision is not local: whether a pixel deserves budget depends on where the rest of the face is, so the representation must let distant positions inform one another. It is also not abstract: the decision has to come back out at full resolution, aligned to the pixels it governs. These two requirements pull in opposite directions, and the architecture is the compromise between them. An encoder reduces the image to a sequence short enough that global interaction is affordable; a state-space stack performs that interaction; a decoder with skip connections restores the spatial detail the encoder discarded. Fig.~\ref{fig:overview} shows the resulting encoder--SSM--decoder path with its two output heads.

\noindent\textbf{Encoder.}
Four stride-2 blocks take the image to a $16{\times}16$ grid. The depth is set by the sequence length we can afford to reason over globally: one block less leaves $1{,}024$ tokens, four times the interaction cost for a spatial precision the decoder can recover anyway; one block more leaves $64$ tokens, at which a single token spans a quarter of the face and the allocation loses the resolution that makes it useful. Each entry below denotes (spatial~resolution$^2 \times$ channels):
\begin{equation}
(256^2{\times}3) \to (128^2{\times}64) \to (64^2{\times}128) \to (32^2{\times}256) \to (16^2{\times}d),
\end{equation}
where $d{=}256$.
Each block applies Conv, GroupNorm, GELU.
The resulting $16{\times}16$ feature map is flattened into a sequence of $N{=}256$ tokens, each representing a $16{\times}16$ spatial region of the input face.

Structured state space models~\citep{gu2021s4} map a scalar input signal to an output through a latent state $\mathbf{h} \in \mathbb{R}^{N'}$ evolving under a linear system $\mathbf{h}' = \mathbf{A}\mathbf{h} + \mathbf{b}u$, $y = \mathbf{c}^\top\mathbf{h}$, applied independently to each of the $d_{\text{model}}$ channels of a token. Mamba~\citep{gu2023mamba} makes $\mathbf{B}_t$, $\mathbf{C}_t$ and the step size $\Delta_t$ linear functions of the full $D$-dimensional token $\mathbf{x}_t$, so that information routing becomes content-dependent, and initializes the state matrix as $\mathbf{A} = -\mathrm{diag}(a_1, \ldots, a_{N'})$ with $a_i > 0$, which reduces its action to element-wise scaling. Discretization gives $\bar{\mathbf{a}}_t = \exp(\Delta_t\, \mathbf{a})$, where $\mathbf{a}$ collects the diagonal entries of $\mathbf{A}$ and $\exp(\cdot)$ is applied element-wise. For the input matrix we use the first-order approximation $\bar{\mathbf{B}}_t = \Delta_t\, \mathbf{B}_t$ adopted by the reference Mamba implementation rather than the exact zero-order-hold expression $(\Delta_t\mathbf{A})^{-1}(\exp(\Delta_t\mathbf{A})-\mathbf{I})\Delta_t\mathbf{B}_t$; the two agree to first order in $\Delta_t$ and the approximation is what the released kernels compute. Writing $x_t^{(c)}$ for the scalar input of channel $c$, the resulting time-variant recurrence is
\begin{equation}
\mathbf{h}_t^{(c)} = \bar{\mathbf{a}}_t \odot \mathbf{h}_{t-1}^{(c)} + \bar{\mathbf{B}}_t\, x_t^{(c)}, \quad
y_t^{(c)} = \mathbf{C}_t^\top \mathbf{h}_t^{(c)} + D_{\text{skip}}\, x_t^{(c)},
\label{eq:ssm_recurrence}
\end{equation}
where $\mathbf{h}_t^{(c)} \in \mathbb{R}^{N'}$ is the per-channel hidden state, $\odot$ the Hadamard product, and $D_{\text{skip}} \in \mathbb{R}$ a learnable feed-through parameter; the per-channel outputs are concatenated into the output token $\mathbf{y}_t \in \mathbb{R}^{D}$.

The full Mamba block wraps the S6 layer within a gated architecture~\citep{gu2023mamba}: the input token is linearly projected and split into two equal-dimensional branches; one branch is processed by a causal 1D convolution followed by the S6 recurrence in Eq.~\eqref{eq:ssm_recurrence}, and its output is multiplied element-wise with the other branch (a gating mechanism) before a final linear projection.

Eq.~\eqref{eq:ssm_recurrence} is the standard Mamba formulation~\citep{gu2023mamba}; we adopt it unchanged. What is specific to this work is the quantity we ask it to produce. The selective update is a mechanism for deciding, per position, how much of the accumulated context to retain, which is structurally the same decision an allocation map encodes; using the recurrence to emit a budget rather than a signal therefore matches the operator to the task instead of borrowing it as a generic feature extractor.

In our implementation, we stack 4 Mamba blocks with $d_{\text{model}}{=}256$ and $d_{\text{state}}{=}16$. The scan is unidirectional in row-major raster order, so a token aggregates context from preceding positions rather than from the entire grid; we return to bidirectional scanning in Section~\ref{sec:discussion_limitations}.

The decoder mirrors the encoder with four transposed convolutional blocks and skip connections from corresponding encoder levels.
Two parallel heads produce the final outputs: the \emph{perturbation head} outputs $\mathbf{p} = \tanh(\cdot) \in [-1, 1]^{3 \times H \times W}$ encoding perturbation direction and relative magnitude, and the \emph{allocation head} outputs $\mathbf{m} = \sigma(\cdot) \in [0, 1]^{1 \times H \times W}$ providing spatial selectivity. The two heads exist because the allocation alone cannot be trained: a budget map has no gradient signal of its own, and it only becomes learnable when something is spent through it. The perturbation head supplies that something during training. At protection time it is no longer needed for this purpose, which is why the composition below is used for learning but not for deployment.
Since $\mathbf{p}$ has three channels and $\mathbf{m}$ has one, we replicate $\mathbf{m}$ along the channel axis to obtain $\tilde{\mathbf{m}} \in [0,1]^{3 \times H \times W}$, where each channel of $\tilde{\mathbf{m}}$ is identical to $\mathbf{m}$.
The final perturbation is then computed as:
\begin{equation}
\boldsymbol{\delta} = \alpha \cdot \epsilon \cdot \bigl(\mathbf{p} \odot \tilde{\mathbf{m}}\bigr),
\label{eq:dual_head}
\end{equation}
where $\alpha \in \mathbb{R}$ is a learnable scalar (initialized near 0.95) and $\odot$ denotes the Hadamard product between two tensors of identical dimension $3 \times H \times W$.

\noindent\textbf{Budget allocation.}
We find the direct composition in Eq.~\eqref{eq:dual_head} counter-productive
(Section~\ref{subsec:ablation}): a perturbation synthesized by the decoder
carries image-independent structure that survives refinement and degrades
distributional realism. We therefore use the mask only to \emph{allocate} the
budget. Let $\tilde{\mathbf{m}} = \min(1,\, 2\,\mathbf{m}/\max_i m_i)$;
the per-pixel budget map is
\begin{equation}
\boldsymbol{\epsilon}
  = \min\!\bigl(\gamma\,\epsilon\,\tilde{\mathbf{m}},\; \epsilon_{\max}\bigr),
\label{eq:budget_map}
\end{equation}
where $\gamma \ge 1$ is the \emph{allocation gain} and $\epsilon_{\max}=0.10$ a
hard global cap. Every projection during refinement is applied element-wise,
\begin{equation}
\Pi_{\boldsymbol{\epsilon}}(\boldsymbol{\delta})_i
  = \max\bigl(\min(\delta_i,\, \epsilon_i),\, -\epsilon_i\bigr),
\label{eq:proj_alloc}
\end{equation}
so $\|\boldsymbol{\delta}\|_\infty \le \epsilon_{\max}$ holds by construction.
The perturbation is initialized at $\boldsymbol{\delta} = \mathbf{0}$ and its
values come entirely from the projected optimization: the network contributes
the allocation, not the noise. Because the learned mask is close to binary
(Section~\ref{subsec:ablation}), increasing $\gamma$ saturates quickly; we use
$\gamma = 1.5$ for all reported results, a mean budget of $0.038$, i.e.,
$42\%$ of the mean per-pixel $\ell_\infty$ capacity a uniform map would allocate at the same cap. The
ablation rows in group~(A) of Table~\ref{tab:ablation} use $\gamma = 1.0$ so
that the effect of the budget \emph{map} is isolated from that of the gain.
\subsection{Loss Functions}
\label{subsec:losses}
We combine identity disruption and visual quality preservation through multiple complementary losses.

\noindent\textbf{Identity loss} $\mathcal{L}_{\text{id}}$ is defined in Eq.~\eqref{eq:lid} and serves as the primary attack objective.

\noindent\textbf{SSIM loss} preserves structural similarity~\citep{wang2004ssim}: $\mathcal{L}_{\text{SSIM}} = 1 - \text{SSIM}(\mathbf{x}, \mathbf{x}_{\text{adv}})$. \textbf{Total variation loss} smooths the perturbation, $\mathcal{L}_{\text{TV}} = \tfrac{1}{3HW}\sum_{c,i,j} |\delta_{c,i+1,j} - \delta_{c,i,j}| + |\delta_{c,i,j+1} - \delta_{c,i,j}|$, with $c$ indexing color channels and $(i,j)$ pixel positions.

\noindent\textbf{Perceptual loss} preserves high-level appearance. A perceptual loss is defined by the feature space it is measured in rather than by the recency of the network that supplies it. Johnson et~al.~\citep{johnson2016perceptual} established the relu activations of VGG-16~\citep{simonyan2015vgg} as that space; the learned perceptual metric of Zhang et~al.~\citep{zhang2018lpips} recalibrates activations of the same class of frozen ImageNet networks with weights fitted to human judgments, and is the perceptual term of \mbox{AMT-GAN}~\citep{huang2022amtgan} (Eq.~11 therein) and of DiffAM~\citep{sun2024diffam} (Eqs.~4 and~10 therein). We use the former, which needs no additional learned layer; a backbone from outside this class would change what the term measures. We use only the first ten convolutional layers, up to \texttt{relu4\_3}, which is $7.64$M of VGG-16's parameters rather than the full network, and they are frozen throughout: gradients flow to $\boldsymbol{\delta}$, never to $\phi$.
The term is $\mathcal{L}_{\text{perc}} = \|\phi(\mathbf{x}) - \phi(\mathbf{x}_{\text{adv}})\|_1$, with $\phi(\cdot)$ the truncated feature stack.

\noindent\textbf{Allocation-specific terms.}
Two terms shape the mask into a usable allocation. A \emph{coverage} term
prevents the degenerate solution in which the network predicts a vanishing
budget everywhere:
\begin{equation}
\mathcal{L}_{\text{cov}}
  = \mathrm{ReLU}\!\left(\kappa_{\max}
      - \tfrac{\|\boldsymbol{\delta}\|_\infty}{\epsilon}\right)
  + \omega\,\mathrm{ReLU}\!\left(\kappa_{\mu}
      - \tfrac{\|\boldsymbol{\delta}\|_1}{3HW\epsilon}\right),
\label{eq:coverage}
\end{equation}
with $\kappa_{\max}{=}0.9$, $\kappa_{\mu}{=}0.2$ and $\omega{=}3$. The second
term is essential: with only the first, the network satisfies the constraint by
placing a few isolated spikes at the cap and leaving the remainder at zero. A
\emph{selectivity} term $\mathcal{L}_{\text{sel}} = \mathbb{E}[m(1-m)]$ drives
the mask toward a binary decision, so that it commits to specific regions rather
than converging to a spatially flat map.

\noindent\textbf{Naturalness loss.} None of the terms above refers to what real faces look like. SSIM and PSNR compare the protected image with its own clean original, and total variation is a smoothness prior on $\boldsymbol{\delta}$ alone; all three are therefore blind to the artifact that matters most here: perturbation that is small everywhere, smooth everywhere, faithful to the original, and still does not look like anything a camera produces. That failure is distributional, and it needs a term that compares against the statistics of real face patches rather than against one reference image. We therefore use binary cross-entropy (BCE) with a lightweight PatchGAN discriminator~\citep{isola2017image} $D$ operating on image patches:
\begin{equation}
\mathcal{L}_{\text{nat}} = \text{BCE}(D(\mathbf{x}_{\text{adv}}),\, \mathbf{1}),
\label{eq:nat_loss}
\end{equation}
which is activated in late training once ASR is stable, and again in Stage~2 of the protected-image refinement (Section~\ref{subsec:optimization}), where $D$ is frozen and contributes only a gradient.
\subsection{Two-Stage Projected Refinement}
\label{subsec:optimization}
Under a strict $\ell_\infty$ cap the two objectives are not merely in tension, they are in tension \emph{asymmetrically}. Identity suppression is hard to reach and easy to lose: a perturbation that has crossed the verification threshold sits close to it, so weighting quality from the start prevents the attack from ever being established, while weighting it after the attack exists mostly removes perturbation the attack was not using. Optimizing both at once therefore settles at a worse point than optimizing them in order. We split the budget into a stage that establishes the attack and a stage that recovers quality under a guard that stops the attack from being given back, and combine this with projected refinement drawing on the transfer attack literature~\citep{dong2018boosting,dong2019evading,xie2019difgsm}.

\noindent\textbf{Stage 1 (attack establishment).}
Iterations $0$ to $T_1$ minimize $\mathcal{L}_{S_1} = \lambda_{\text{id}}^{(1)}\mathcal{L}_{\text{id}} + \lambda_{\text{s}}^{(1)}\mathcal{L}_{\text{SSIM}} + \lambda_{\text{tv}}^{(1)}\mathcal{L}_{\text{TV}}$ with $\lambda_{\text{id}}^{(1)}{=}14$, $\lambda_{\text{s}}^{(1)}{=}2$ and $\lambda_{\text{tv}}^{(1)}{=}0.3$, to reach a strong ASR basin quickly; the projection uses the full per-pixel budget map from the first iteration.

\noindent\textbf{Stage 2 (quality recovery with ASR Lock).}
Iterations $T_1{+}1$ to $T_2$ shift the emphasis to quality and add the perceptual and naturalness terms:
\begin{equation}
\mathcal{L}_{S_2} = \lambda_{\text{id}}^{(2)} \mathcal{L}_{\text{id}} + \lambda_{\text{s}}^{(2)} \mathcal{L}_{\text{SSIM}} + 0.3\,\mathcal{L}_{\text{TV}} + 2\,\mathcal{L}_{\text{perc}} + \lambda_{\text{nat}} \mathcal{L}_{\text{nat}},
\label{eq:stage2}
\end{equation}
where the remaining three weights are set by the \emph{ASR Lock} from the surrogate success rate $\widehat{\text{ASR}}_t$, the fraction of image--surrogate pairs whose similarity to the clean embedding is below $\tau$, re-estimated every 10 iterations:
\begin{equation}
\renewcommand{\arraystretch}{0.62}
\bigl(\lambda_{\text{id}}^{(2)},\, \lambda_{\text{s}}^{(2)},\, \lambda_{\text{nat}}\bigr) =
\left\{\begin{array}{@{}l@{\quad}l@{}}
(12,\, 4,\, 1) & \widehat{\text{ASR}}_t < 0.80, \\
(10,\, 6,\, 2) & 0.80 \le \widehat{\text{ASR}}_t < 0.85, \\
(6,\, 10,\, 3) & \widehat{\text{ASR}}_t \ge 0.85.
\end{array}\right.
\label{eq:asr_lock}
\end{equation}
Weight moves from identity to quality only while the surrogate attack holds and moves back as soon as it weakens. If the final $\widehat{\text{ASR}}$ ends more than 5 points below the best value observed, the perturbation from that best check is returned.

During both stages, $\boldsymbol{\delta}$ is refined via a short projected optimization that integrates momentum~\citep{dong2018boosting}, translation-invariant (TI) gradient smoothing~\citep{dong2019evading}, and input diversity (DI, following the convention of Xie et~al.~\citep{xie2019difgsm}) to improve transferability.
Let $\mathcal{T}(\cdot)$ be a random resizing-and-padding transformation~\citep{xie2019difgsm} applied with probability $p{=}0.5$ in Stage~1 and $p{=}0.3$ in Stage~2, and $\mathbf{K}$ a Gaussian kernel for TI smoothing.
At iteration $t$:
\begin{align}
\tilde{\mathbf{x}}_{\text{adv}}^{(t)} &= \mathcal{T}\!\left(\mathrm{clip}(\mathbf{x} + \boldsymbol{\delta}^{(t)}, 0, 1)\right), \label{eq:di} \\
\mathbf{g}_t &= \mathbf{K} * \nabla_{\boldsymbol{\delta}}\mathcal{L}\!\left(\tilde{\mathbf{x}}_{\text{adv}}^{(t)}\right), \label{eq:ti} \\
\mathbf{v}_t &= \mu\, \mathbf{v}_{t-1} + \frac{\mathbf{g}_t}{\|\mathbf{g}_t\|_1 + \eta}, \label{eq:momentum} \\
\boldsymbol{\delta}^{(t+1)} &= \Pi_{\boldsymbol{\epsilon}^{(t)}}\!\left(\boldsymbol{\delta}^{(t)} - \alpha_{\text{s}}\, \mathrm{sign}(\mathbf{v}_t)\right), \label{eq:project}
\end{align}
where the division is element-wise, $\mu{=}0.9$ and $\eta{=}10^{-8}$ prevents division by zero. The TI kernel $\mathbf{K}$ is $15{\times}15$ with $\sigma{=}3$. The step size is tied to the budget and the iteration count rather than tuned separately, $\alpha_{\text{s}}{=}2\epsilon_t/T$ with $T{=}200$ total iterations ($110$ in Stage~1, $90$ in Stage~2). The nominal budget is scheduled, and the schedule enters the projection, not only the step size. The map is rebuilt each iteration as $\boldsymbol{\epsilon}^{(t)} = \min(\gamma\,\epsilon_t\,\tilde{\mathbf{m}},\, \epsilon_{\max})$, with $\epsilon_t$ held at $1.05\,\epsilon$ through Stage~1 and annealed to $0.90\,\epsilon$ across Stage~2. The cap applies to every rebuilt map, so the schedule redistributes capacity over time and never relaxes the bound: $\|\boldsymbol{\delta}\|_\infty \le \epsilon_{\max}$ holds at every iteration.

\noindent\textbf{GIL.}
To escape local minima and encourage semantically consistent perturbation patterns, every $M$ refinement steps we inject a generator-guided correction.
The generator is budget-conditioned: the scalar budget enters as the multiplier of the dual-head composition in Eq.~\eqref{eq:dual_head}, so $G(\mathbf{x};\epsilon)$ denotes the same network evaluated with that multiplier and $G(\mathbf{x})$ is shorthand for the nominal setting. Here it re-processes the current adversarial image with a reduced budget $\epsilon'{=}0.3\,\epsilon$:
\begin{equation}
(\mathbf{p},\, \mathbf{m}) = G(\mathbf{x}_{\text{adv}};\, \epsilon'), \quad
\boldsymbol{\delta} \leftarrow \Pi_{\boldsymbol{\epsilon}^{(t)}}\!\left(\boldsymbol{\delta} + \beta\, (\mathbf{p} \odot \tilde{\mathbf{m}})\right),
\label{eq:gil}
\end{equation}
where $\beta \in \mathbb{R}$ is a scalar controlling injection strength ($\beta{=}0.25$ in our experiments) and $\tilde{\mathbf{m}}$ is the channel-replicated mask as in Eq.~\eqref{eq:dual_head}.
GIL is applied every $M_1{=}15$ iterations in Stage~1 and every $M_2{=}25$ in Stage~2; because
the injection is immediately re-projected onto the budget map, its effect is
confined to the allocated regions.

\noindent\textbf{Frequency suppression.}
The sign-based update of Eq.~\eqref{eq:project} moves every pixel by the same amount at every step, which concentrates the perturbation's spectrum in the highest frequencies the grid can represent, exactly where the eye is most sensitive to texture that does not belong. Attenuating that band costs little attack strength because identity evidence is not carried there. Every 15 iterations in the later part of Stage~2 we therefore smooth
$\boldsymbol{\delta}$ before re-projection. A $2{\times}2$ Haar-style analysis
splits $\boldsymbol{\delta}$ into one low-pass and three band-pass components;
the diagonal (HH) component is attenuated by a factor $0.75$ and the components
are recombined by nearest-neighbour upsampling. The operator is therefore a
band-attenuating smoother rather than an exact wavelet reconstruction: it
suppresses high-frequency structure within each $2{\times}2$ neighbourhood,
which removes perceptually salient artifacts at negligible cost to attack
performance.
\subsection{Training Protocol}
\label{subsec:training}
We train the generator $G$ end-to-end with a curriculum schedule designed to stabilize optimization under the bounded-perturbation constraint. \emph{Phase~1 (Escape)}: the identity loss is active from the first epoch together with the coverage term of Eq.~\eqref{eq:coverage}, establishing a non-degenerate allocation; a warm-up driven purely by quality objectives instead collapses the network to a vanishing output, from which the subsequent attack phase does not recover.

\emph{Phase~2 (Attack strengthening)}: the identity weight is raised further while coverage pressure is maintained, transitioning into the two-stage refinement described in Sec.~\ref{subsec:optimization}.

\emph{Phase~3 (Late-stage stabilization)}: ASR Lock activates, the coverage pressure is annealed toward zero so the network can withdraw budget from regions that do not need it, and the PatchGAN discriminator~\citep{isola2017image} provides naturalness regularization.

Optimizer settings, the surrogate ensemble and the hardware are given in Section~\ref{subsec:exp_setup}. The complete pipeline is summarized in Algorithm~\ref{alg:ssm_attack}.

\begin{algorithm}[t]
\caption{SSM-Attack: Mamba-Based Black-Box Face Protection}
\label{alg:ssm_attack}
{\fontsize{9}{9}\selectfont
\begin{algorithmic}[1]
\Require Face image $\mathbf{x}$, surrogates $\{f_k\}_{k=1}^{K}$, allocation network $G$, nominal budget $\epsilon$, cap $\epsilon_{\max}$, gain $\gamma$
\Ensure Protected image $\mathbf{x}_{\text{adv}}$
\State \textbf{Allocation shape (single forward pass):}
\State $(\cdot,\, \mathbf{m}) \gets G(\mathbf{x})$;\quad $\tilde{\mathbf{m}} \gets \min(1,\, 2\,\mathbf{m}/\max_i m_i)$
\State $\boldsymbol{\delta} \gets \mathbf{0}$
\State \textbf{Stage 1 (Attack Establishment):}
\For{$t = 1$ to $T_1$}
    \State $\epsilon_t \gets 1.05\,\epsilon$;\quad $\boldsymbol{\epsilon}^{(t)} \gets \min(\gamma\,\epsilon_t\,\tilde{\mathbf{m}},\, \epsilon_{\max})$
    \State $\mathbf{x}_{\text{adv}} \gets \mathrm{clip}(\mathbf{x} + \boldsymbol{\delta}, 0, 1)$
    \State Compute $\mathcal{L}_{\text{id}}$ on surrogate ensemble $\{f_k\}$
    \State Update $\boldsymbol{\delta}$ with DI, TI and momentum; project element-wise via $\Pi_{\boldsymbol{\epsilon}^{(t)}}$ (Eqs.~\eqref{eq:di}--\eqref{eq:project})
    \If{$t \bmod M_1 = 0$}
        \State Apply GIL correction, re-projected by $\Pi_{\boldsymbol{\epsilon}^{(t)}}$ (Eq.~\eqref{eq:gil})
    \EndIf
\EndFor
\State \textbf{Stage 2 (Quality Recovery + ASR Lock):}
\For{$t = T_1+1$ to $T_2$}
    \State $\epsilon_t \gets \epsilon\,\bigl(1 - 0.1\,\tfrac{t-T_1}{T_2-T_1}\bigr)$;\quad $\boldsymbol{\epsilon}^{(t)} \gets \min(\gamma\,\epsilon_t\,\tilde{\mathbf{m}},\, \epsilon_{\max})$
    \State $\mathbf{x}_{\text{adv}} \gets \mathrm{clip}(\mathbf{x} + \boldsymbol{\delta}, 0, 1)$
    \State Estimate $\widehat{\text{ASR}}_t$; adapt weights (Eq.~\eqref{eq:asr_lock})
    \State Update $\boldsymbol{\delta}$ with quality-dominant losses (Eq.~\eqref{eq:stage2}), including $\mathcal{L}_{\text{nat}}$ from the frozen discriminator
    \State Apply wavelet HH suppression, then $\Pi_{\boldsymbol{\epsilon}^{(t)}}$
    \If{$t \bmod M_2 = 0$}
        \State Apply GIL correction, re-projected by $\Pi_{\boldsymbol{\epsilon}^{(t)}}$ (Eq.~\eqref{eq:gil})
    \EndIf
\EndFor
\State \Return $\mathbf{x}_{\text{adv}} \gets \mathrm{clip}(\mathbf{x} + \boldsymbol{\delta}, 0, 1)$
\end{algorithmic}}
\end{algorithm}

\noindent\textbf{Computational complexity.} Mamba processes $N{=}256$ spatial tokens with $O(N \cdot d_{\text{model}} \cdot d_{\text{state}})$ complexity per block, compared to $O(N^2 \cdot d_{\text{model}})$ for self-attention in Transformers~\citep{dosovitskiy2021vit}. Combined with a compact 5.73M-parameter allocation network, this linear-time formulation avoids both the quadratic cost of attention and the heavy pretrained backbones (100M--1B+ parameters) required by diffusion- and portrait-based pipelines.
\section{Experiments}
\label{sec:experiments}
\subsection{Experimental Setup}
\label{subsec:exp_setup} We adopt the two widely used benchmarks in face privacy protection research. From \textbf{CelebA-HQ}~\citep{karras2018celeba} we take the 1{,}000-image subset of distinct identities used by AMT-GAN~\citep{huang2022amtgan}, DiffAM~\citep{sun2024diffam} and Adv-CPG~\citep{wang2025advcpg}. Those methods split the subset into groups and redirect each toward a chosen target identity, and their reported ASR counts impersonation, $\cos(f(\mathbf{x}_{\text{adv}}), f(\mathbf{x}_{\text{tgt}})) > \tau$ for a target image $\mathbf{x}_{\text{tgt}}$. SSM-Attack instead performs \textbf{untargeted identity dodging}, minimizing Eq.~\eqref{eq:lid} with no destination identity, over all 1{,}000 images under one protocol, and its ASR counts $\cos(f(\mathbf{x}), f(\mathbf{x}_{\text{adv}})) < \tau$. The two rates answer different questions. Dodging is generally the easier objective, so part of our margin may reflect the task rather than attack strength, while impersonation success overstates protection, since an example that passes as the target can still match its source~\citep{zhou2024rethinking}. We report dodging because it is the quantity privacy protection requires.

LADN~\citep{gu2019ladn} is a makeup transfer dataset comprising 333 non-makeup and 302 makeup images. We use 332 non-makeup images as the test set, processed with the same untargeted protocol and identical hyperparameters.

The allocation network is trained on the Makeup Transfer (MT) dataset~\citep{li2018beautygan} (1{,}115 non-makeup and 2{,}719 makeup images), which AMT-GAN~\citep{huang2022amtgan} and DiffAM~\citep{sun2024diffam} also use for training, combined with face images from one of CelebA-HQ or FFHQ. Which combination produces which result is stated in Section~\ref{sec:discussion_limitations}; in every case the evaluation set is disjoint from the training data.

All face images are aligned and cropped using the multi-task cascaded convolutional network detector~\citep{zhang2016mtcnn} to $112 \times 112$ for IR-152, IRSE-50, and MobileFace (MF), and to $160 \times 160$ for FaceNet following its native input specification~\citep{schroff2015facenet}, and resized to $256 \times 256$ for perturbation generation.

The four target models span three training objectives and three backbone families. \textbf{IR-152} and \textbf{IRSE-50}~\citep{deng2019arcface} are residual networks trained with ArcFace loss, the second adding Squeeze-and-Excitation blocks. \textbf{MobileFace}~\citep{chen2018mobilefacenets} is a depthwise-separable network trained with Softmax loss. \textbf{FaceNet}~\citep{schroff2015facenet} differs on both axes: an InceptionResNet-v1 backbone trained with triplet loss on VGGFace2, which is why it is the hardest target for surrogates drawn from the other three. Verification thresholds $\tau$ follow standard practice~\citep{huang2022amtgan,sun2024diffam,wang2025advcpg}: $0.167$, $0.241$, $0.409$ and $0.302$ respectively, each at FAR@0.01.

We employ the LOO evaluation protocol standard in this literature~\citep{yang2021tipim,huang2022amtgan,sun2024diffam,wang2025advcpg}: for each target $f_t$ the perturbation is optimized against the remaining three models as a surrogate ensemble and scored only on $f_t$, repeated for all four hold-out choices.

\noindent\textbf{The leave-one-out rule also governs training.} A separate allocation network is trained for each held-out target, supervised only by that configuration's three surrogates; the held-out model is seen neither at training nor at protection time. Four networks are therefore trained per image corpus; Table~\ref{tab:protocol} states the full assignment.

\begin{table}[t]
\centering
\caption{\textbf{Protocol assignment.} One allocation network per held-out target per image corpus; the held-out model supervises neither training nor refinement. MT is the Makeup Transfer corpus~\citep{li2018beautygan}. Evaluation images are disjoint from the images used to train the corresponding network in every row.}
\label{tab:protocol}
\setlength{\tabcolsep}{5pt}
\renewcommand{\arraystretch}{1.08}
\begin{tabular}{ll|l|l}
\toprule
Test set & \shortstack[l]{Held-out\\target} & \shortstack[l]{Surrogates\\(training \emph{and} refinement)} & \shortstack[l]{Images\\for $G$} \\
\midrule
\multirow{4}{*}{\shortstack[l]{CelebA-HQ\\LADN}}
 & IR-152     & IRSE-50, FaceNet, MobileFace & \multirow{4}{*}{FFHQ $+$ MT} \\
 & IRSE-50    & IR-152, FaceNet, MobileFace  & \\
 & FaceNet    & IR-152, IRSE-50, MobileFace  & \\
 & MobileFace & IR-152, IRSE-50, FaceNet     & \\
\midrule
\multirow{4}{*}{FFHQ}
 & IR-152     & IRSE-50, FaceNet, MobileFace & \multirow{4}{*}{CelebA-HQ $+$ MT} \\
 & IRSE-50    & IR-152, FaceNet, MobileFace  & \\
 & FaceNet    & IR-152, IRSE-50, MobileFace  & \\
 & MobileFace & IR-152, IRSE-50, FaceNet     & \\
\bottomrule
\end{tabular}
\end{table}

Attack effectiveness is reported as ASR~\citep{sharif2019general} at FAR@0.01 under face verification, for each of the four held-out targets and as their mean in the ``Avg.'' column. Fidelity is reported with SSIM~\citep{wang2004ssim} and PSNR in decibels, which compare a protected image with its own original, and FID~\citep{heusel2017fid}, which compares the protected and clean \emph{sets} in a deep feature space. The first two are the primary indicators for $\ell_\infty$-bounded noise; FID is the more informative one for unrestricted methods that alter facial semantics. ASR is a proportion over the evaluation set, so its binomial standard error is about $1.1$ points at the reported CelebA-HQ and FFHQ accuracies over $1{,}000$ images and about $1.9$ points on the $332$-image LADN set; differences below roughly two points should be read accordingly.

We compare against 18 published baselines in four categories --- five noise-based, seven makeup-based, five facial-semantic-invariant and one portrait-based, named individually in Tables~\ref{tab:asr_celeba_ladn} and~\ref{tab:ffhq}~\citep{goodfellow2015explaining,dong2018boosting,madry2018pgd,dong2019evading,yang2021tipim,komkov2021advhat,yin2021advmakeup,huang2022amtgan,shamshad2023clip2protect,li2024gift,shamshad2024dfpp,sun2024diffam,liu2025diffprotect,zhang2024dpg,an2024sd4privacy,liu2024advdiffusion,chow2025p3mask,wang2025advcpg}.

Rather than mixing self-reported values obtained under differing protocols, we take their numbers from unified benchmark tables: Adv-CPG~\citep{wang2025advcpg} for CelebA-HQ and FFHQ, DiffAM~\citep{sun2024diffam} for LADN. Methods absent from a benchmark are marked ``--''.

The generator of Section~\ref{subsec:generator} holds 5.73M parameters ($d_{\text{model}}{=}256$, $d_{\text{state}}{=}16$). Training follows the three-phase curriculum of Section~\ref{subsec:training} with AdamW~\citep{loshchilov2019adamw}, learning rates $2\times10^{-4}$ for the generator and $1\times10^{-4}$ for the discriminator, weight decay $0.01$, cosine annealing and gradient clipping at $1.0$.

At protection time a single forward pass produces the budget map, and a two-stage refinement then fills in the values. Stage~1 runs for iterations 0--109 and Stage~2 for 110--199 with the weights of Section~\ref{subsec:optimization}; Stage~2 adds, under ASR~Lock, the perceptual term, the wavelet suppression and the naturalness gradient of the frozen discriminator~\citep{isola2017image}. GIL fires every 15 iterations in Stage~1 and every 25 in Stage~2.

We adopt a \textbf{model-adaptive perturbation budget} under $\ell_\infty$ norm to account for the varying perturbation sensitivity of different FR architectures:
$\epsilon = 0.09$ ($\approx$23.0/255) for IR-152 and MobileFace, $\epsilon = 0.095$ ($\approx$24.2/255) for IRSE-50, and $\epsilon = 0.10$ ($\approx$25.5/255) for FaceNet.
All budgets satisfy $\epsilon \leq 0.10$ and are enforced by the three-fold projection described in Section~\ref{subsec:problem}. Unless otherwise stated, all settings are identical across both datasets.
Experiments are conducted on a single NVIDIA T4 GPU.
\subsection{Comparison on Black-Box Attacks}
\label{subsec:main_results}
\begin{table*}[t]
\centering
\caption{\textbf{Black-box ASR (\%)} on CelebA-HQ and LADN. FFHQ is Table~\ref{tab:ffhq}.
CelebA-HQ baselines follow Adv-CPG~\citep{wang2025advcpg} Table~2, LADN
baselines DiffAM~\citep{sun2024diffam} Table~1, and where both report a method
we quote the higher value. $^{\ddagger}$Adv-CPG's LADN results come from its
supplementary Table~7. ``--'' marks methods absent from those tables; MF is
MobileFace. \textbf{Bold}: best. \underline{Underline}: second.}
\label{tab:asr_celeba_ladn}
\setlength{\tabcolsep}{1.2pt}
\renewcommand{\arraystretch}{0.85}
\resizebox{\textwidth}{!}{%
\begin{tabular}{ll|ccccc|ccccc}
\toprule
\multirow{2}{*}{Category} & \multirow{2}{*}{Method}
& \multicolumn{5}{c|}{CelebA-HQ}
& \multicolumn{5}{c}{LADN} \\
\cmidrule(lr){3-7}\cmidrule(lr){8-12}
& & IR-152 & IRSE-50 & FaceNet & MF & Avg.
  & IR-152 & IRSE-50 & FaceNet & MF & Avg. \\
\midrule
-- & Clean
& 4.64 & 5.74 & 1.06 & 13.27 & 6.18
& 3.61 & 2.71 & 0.60 & 5.11 & 3.01 \\
\midrule
\multirow{5}{*}{Noise}
& FGSM~\citep{goodfellow2015explaining}
& 12.07 & 45.81 & 1.35 & 53.02 & 28.06
& -- & -- & -- & -- & -- \\
& MI-FGSM~\citep{dong2018boosting}
& 46.53 & 70.57 & 27.09 & 58.94 & 50.78
& 25.57 & 48.90 & 6.31 & 45.01 & 31.45 \\
& PGD~\citep{madry2018pgd}
& 41.87 & 63.23 & 19.62 & 57.34 & 45.52
& 19.59 & 40.09 & 3.82 & 41.09 & 26.15 \\
& TI-DIM~\citep{dong2019evading}
& 35.13 & 62.38 & 13.68 & 52.84 & 41.01
& 34.18 & 56.36 & 22.11 & 48.30 & 40.24 \\
& TIP-IM~\citep{yang2021tipim}
& 41.26 & 57.29 & 39.07 & 49.56 & 46.80
& 43.57 & 65.89 & 63.50 & 46.48 & 54.86 \\
\midrule
\multirow{7}{*}{Makeup}
& Adv-Hat~\citep{komkov2021advhat}
& 5.04 & 16.88 & 4.91 & 12.64 & 9.87
& -- & -- & -- & -- & -- \\
& Adv-Makeup~\citep{yin2021advmakeup}
& 12.68 & 19.95 & 1.37 & 22.11 & 14.03
& 10.03 & 29.64 & 0.97 & 22.38 & 15.76 \\
& AMT-GAN~\citep{huang2022amtgan}
& 12.09 & 53.26 & 4.87 & 47.95 & 29.54
& 49.12 & 89.64 & 32.13 & 72.43 & 60.83 \\
& CLIP2Protect~\citep{shamshad2023clip2protect}
& 47.63 & 80.96 & 42.57 & 73.64 & 61.20
& 53.31 & 91.57 & 47.91 & 79.94 & 68.18 \\
& GIFT~\citep{li2024gift}
& 73.84 & 83.72 & 56.48 & \underline{86.37} & 75.10
& -- & -- & -- & -- & -- \\
& DFPP~\citep{shamshad2024dfpp}
& 46.38 & 80.59 & 45.37 & 72.13 & 61.12
& -- & -- & -- & -- & -- \\
& DiffAM~\citep{sun2024diffam}
& 65.09 & \textbf{89.66} & 62.99 & 84.51 & 75.56
& 66.75 & \underline{95.66} & 65.44 & \textbf{92.04} & \underline{79.97} \\
\midrule
\multirow{5}{*}{\shortstack{Sem.\\Inv.}}
& DiffProtect~\citep{liu2025diffprotect}
& 58.64 & 79.34 & 24.69 & 75.91 & 59.65
& -- & -- & -- & -- & -- \\
& DPG~\citep{zhang2024dpg}
& 42.89 & 62.47 & 35.83 & 66.42 & 51.90
& -- & -- & -- & -- & -- \\
& SD4Privacy~\citep{an2024sd4privacy}
& 66.89 & 79.96 & 53.49 & 74.58 & 68.73
& -- & -- & -- & -- & -- \\
& Adv-Diffusion~\citep{liu2024advdiffusion}
& 52.84 & 81.67 & 34.95 & 70.78 & 60.06
& -- & -- & -- & -- & -- \\
& P3-Mask~\citep{chow2025p3mask}
& 73.47 & 83.40 & 60.24 & 69.64 & 71.69
& -- & -- & -- & -- & -- \\
\midrule
Portrait
& Adv-CPG~\citep{wang2025advcpg}
& \underline{76.96} & \underline{88.72} & \underline{63.50} & \textbf{87.95} & \underline{79.28}
& \underline{70.56} & 91.46 & \textbf{70.42} & 79.54 & 78.00$^{\ddagger}$ \\
\midrule
SSM
& \textbf{SSM-Attack (Ours)}
& \textbf{95.20} & 87.60 & \textbf{80.40} & 85.20 & \textbf{87.10}
& \textbf{93.07} & \textbf{96.99} & \underline{67.47} & \underline{90.06} & \textbf{86.90} \\
\bottomrule
\end{tabular}}
\end{table*}
As shown in Table~\ref{tab:asr_celeba_ladn}, on CelebA-HQ SSM-Attack achieves the highest average ASR among all 19 evaluated methods, surpassing the previous best Adv-CPG~\citep{wang2025advcpg} by $+7.82$ points.

The per-model pattern is informative. SSM-Attack reaches the highest ASR of any method on \textbf{IR-152} and on \textbf{FaceNet}, the target generally reported as hardest: its InceptionResNet backbone and triplet-loss objective are the furthest from any of its three surrogates~\citep{schroff2015facenet}. That the largest gain appears where the surrogate--target gap is widest suggests that \emph{where} identity evidence resides transfers more readily than \emph{which} texture directions disrupt a particular embedding. On \textbf{IRSE-50} and \textbf{MobileFace} we remain within roughly two points of the best unrestricted methods while operating under an $\ell_\infty$ bound those methods do not impose.

Against the noise-based baselines the margin in average ASR exceeds $30$ points, and against the makeup-based ones SSM-Attack leads DiffAM and CLIP2Protect on average while requiring neither a reference makeup image nor a text prompt.

\begin{table*}[t]
\centering
\caption{Black-box ASR (\%) on FFHQ (1{,}000 images) under the LOO protocol. Baselines from Adv-CPG~\citep{wang2025advcpg} Table~2; ``--'' marks methods absent there. Our networks are trained on CelebA-HQ$+$MT and never saw FFHQ. MF is MobileFace. \textbf{Bold}: best. \underline{Underline}: second.}
\label{tab:ffhq}
\setlength{\tabcolsep}{4pt}
\renewcommand{\arraystretch}{1.0}
\begin{tabular}{ll|cccc|c}
\toprule
Category & Method & IR-152 & IRSE-50 & FaceNet & MF & Avg. \\
\midrule
-- & Clean & 3.14 & 4.73 & 1.49 & 8.96 & 4.58 \\
\midrule
\multirow{5}{*}{Noise}
& FGSM~\citep{goodfellow2015explaining} & 9.86 & 48.53 & 4.23 & 51.48 & 28.53 \\
& MI-FGSM~\citep{dong2018boosting} & 46.31 & 69.24 & 20.62 & 69.26 & 51.36 \\
& PGD~\citep{madry2018pgd} & 31.64 & 75.38 & 18.59 & 63.12 & 47.18 \\
& TI-DIM~\citep{dong2019evading} & 43.57 & 65.89 & 14.72 & 53.31 & 44.37 \\
& TIP-IM~\citep{yang2021tipim} & 46.25 & 67.38 & 59.82 & 52.03 & 56.37 \\
\midrule
\multirow{4}{*}{Makeup}
& AMT-GAN~\citep{huang2022amtgan} & 11.52 & 56.02 & 9.74 & 41.32 & 29.65 \\
& CLIP2Protect~\citep{shamshad2023clip2protect} & 52.12 & 86.53 & 45.01 & 76.29 & 64.99 \\
& GIFT~\citep{li2024gift} & 69.72 & 87.64 & 54.49 & 82.93 & 73.70 \\
& DiffAM~\citep{sun2024diffam} & 67.24 & \underline{90.30} & \underline{64.96} & \underline{89.56} & 78.02 \\
\midrule
\multirow{4}{*}{\shortstack{Sem.\\Inv.}}
& DiffProtect~\citep{liu2025diffprotect} & 57.62 & 60.14 & 49.38 & 67.52 & 58.67 \\
& SD4Privacy~\citep{an2024sd4privacy} & 51.31 & 79.94 & 43.57 & 71.55 & 61.59 \\
& Adv-Diffusion~\citep{liu2024advdiffusion} & 50.93 & 81.76 & 30.84 & 67.52 & 57.76 \\
& P3-Mask~\citep{chow2025p3mask} & 73.18 & 85.35 & 57.92 & 70.69 & 71.79 \\
\midrule
Portrait & Adv-CPG~\citep{wang2025advcpg} & \underline{75.26} & \textbf{91.03} & 63.84 & \textbf{89.94} & \underline{80.02} \\
\midrule
SSM & \textbf{SSM-Attack (Ours)} & \textbf{92.70} & 79.60 & \textbf{70.90} & 81.40 & \textbf{81.15} \\
\bottomrule
\end{tabular}
\end{table*}

To validate cross-dataset generalization, we evaluate on 1{,}000 FFHQ~\citep{karras2019style} images following the identical protocol (Table~\ref{tab:ffhq}). The network used here is the one trained on CelebA-HQ$+$MT, so FFHQ is unseen both as a dataset and as a set of identities. Even under this stricter condition SSM-Attack attains the highest average ASR ($81.15\%$), ahead of Adv-CPG ($80.02\%$) and DiffAM ($78.02\%$), and ranks first on the two hardest targets, IR-152 ($92.70\%$) and FaceNet ($70.90\%$), evidence that the allocation transfers across architectures and across datasets. The margin is narrower than on CelebA-HQ: on IRSE-50 and MobileFace we trail the strongest unrestricted baselines by roughly ten points, consistent with FFHQ's greater identity diversity and with the strict $\ell_\infty$ bound; we discuss this further in Section~\ref{sec:discussion_limitations}.

On LADN (Table~\ref{tab:asr_celeba_ladn}) SSM-Attack again attains the highest average ASR among the methods with reported results, narrowly ahead of DiffAM~\citep{sun2024diffam}, and the per-model ordering repeats: strongest on IR-152 and IRSE-50, weaker on MobileFace and FaceNet. MobileFace's compact depthwise-separable design responds more to the large-scale semantic edits of makeup-based methods than to bounded noise, and the FaceNet gap is amplified on a 332-image set.

Across the three evaluation sets the average ASR spans $5.95$ points ($87.10$ / $86.90$ / $81.15$) and SSM-Attack ranks first on each, whereas several baselines swing by more than $30$ points between sets (AMT-GAN, for instance). Since each dataset is processed independently (Section~\ref{subsec:exp_setup}), the lower FFHQ figure reflects the difficulty of its more diverse identities under a strict $\ell_\infty$ bound, not a train/test domain shift.
\subsection{Image Quality Evaluation}
\label{subsec:quality}
\begin{figure*}[t]
\centering
\includegraphics[width=\textwidth]{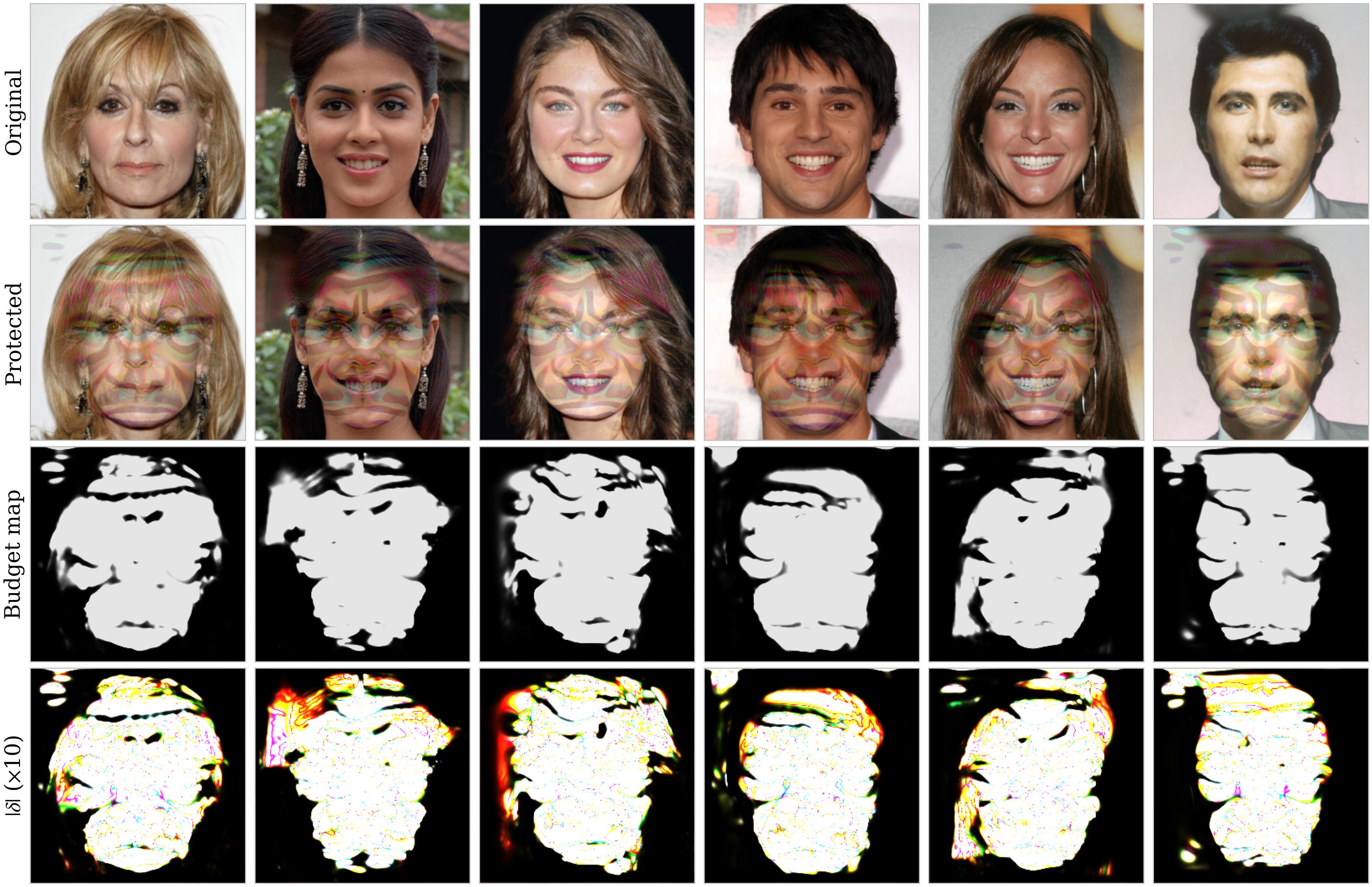}
\caption{\textbf{What the allocation network learns.} Each column is a
CelebA-HQ subject; rows are the original, the protected image, the budget map
$\boldsymbol{\epsilon}$ (brighter is more budget) and $|\boldsymbol{\delta}|$
amplified $10\times$. The budget concentrates on the periocular--nose--mouth
region and the hair and jaw boundaries. No landmark detector or segmentation
mask is used at any point, so this structure is learned; an elliptical prior at
comparable capacity recovers less than half the attack performance
(Table~\ref{tab:ablation}c). Best viewed zoomed in.}
\label{fig:qualitative}
\end{figure*}
Figure~\ref{fig:qualitative} provides a qualitative comparison across all three datasets, confirming that the protected images remain visually close to the originals despite the high attack success rate.

Table~\ref{tab:quality} requires careful interpretation, because noise-based and unrestricted methods operate under fundamentally different quality constraints.

\noindent\textbf{Noise-based comparison.} SSM-Attack achieves SSIM\,=\,0.888 and PSNR\,=\,28.42\,dB on CelebA-HQ under a model-adaptive $\ell_\infty$ perturbation budget ($\epsilon \leq 0.10$). Compared to TIP-IM~\citep{yang2021tipim} (SSIM\,=\,0.924, PSNR\,=\,33.21\,dB), SSM-Attack accepts a moderate fidelity reduction in exchange for a dramatically higher ASR: \textbf{+35.35\%} (87.10\% vs.~46.80\%). Part of that fidelity margin is expected rather than earned: TIP-IM's published protocol adopts $\epsilon = 12/255 \approx 0.047$, about half the cap used here, so the ASR gap is the more informative of the two comparisons. Against the unrestricted methods, SSIM is higher than AMT-GAN's~\citep{huang2022amtgan} ($0.888$ vs.~$0.792$) and DiffAM's~\citep{sun2024diffam} ($0.886$), and PSNR is more than $7$\,dB higher than either ($28.42$ vs.~$19.50$ and $20.53$), which reflects the smaller mean perturbation magnitude allocation permits.

Our FID ($53.6$) is higher than that of the diffusion-based methods (DiffAM $26.10$, Adv-CPG $26.08$), which follows from the paradigms rather than from tuning: semantic edits move an image along the natural image manifold and so stay distributionally aligned, whereas an additive pixel-domain perturbation can shift the feature distribution even when it is barely visible. Adv-Makeup~\citep{yin2021advmakeup} makes the point that quality alone settles nothing: it posts the best scores in the table (SSIM $0.985$, FID $4.23$) by editing only the eye-shadow region, and reaches $14.03\%$ ASR.

\begin{table}[t]
\centering
\caption{Image quality on CelebA-HQ and LADN. CelebA-HQ baselines from Adv-CPG~\citep{wang2025advcpg} Table~4. SSIM and PSNR are the primary indicators for $\ell_\infty$-bounded methods, FID for unrestricted ones. No competing method reports quality metrics on LADN, hence ``--''; our LADN values show that fidelity is stable across datasets. $^{\dagger}$FID is omitted there because the 332-image estimator is biased upward.}
\label{tab:quality}
\setlength{\tabcolsep}{4pt}
\renewcommand{\arraystretch}{1.05}
\begin{tabular}{l|l|ccc|ccc}
\toprule
\multirow{2}{*}{Type} & \multirow{2}{*}{Method}
& \multicolumn{3}{c|}{CelebA-HQ} & \multicolumn{3}{c}{LADN} \\
\cmidrule(lr){3-5}\cmidrule(lr){6-8}
& & SSIM$\uparrow$ & PSNR$\uparrow$ & FID$\downarrow$
  & SSIM$\uparrow$ & PSNR$\uparrow$ & FID$\downarrow$ \\
\midrule
\multirow{2}{*}{Noise}
& TIP-IM~\citep{yang2021tipim}          & 0.924 & 33.21 & 38.73 & -- & -- & -- \\
& \textbf{SSM-Attack (Ours)}            & \textbf{0.888} & \textbf{28.42} & 53.6 & \textbf{0.870} & \textbf{28.10} & --$^{\dagger}$ \\
\midrule
\multirow{4}{*}{Unrest.}
& Adv-Makeup~\citep{yin2021advmakeup}    & 0.985 & 34.52 & 4.23  & -- & -- & -- \\
& AMT-GAN~\citep{huang2022amtgan}        & 0.792 & 19.50 & 34.57 & -- & -- & -- \\
& DiffAM~\citep{sun2024diffam}           & 0.886 & 20.53 & 26.10 & -- & -- & -- \\
& Adv-CPG~\citep{wang2025advcpg}         & 0.898 & 30.00 & 26.08 & -- & -- & -- \\
\bottomrule
\end{tabular}
\end{table}
On LADN, SSM-Attack maintains consistent quality (SSIM\,=\,0.870, PSNR\,=\,28.10\,dB), confirming that quality remains stable across datasets.

\subsection{Computational Cost Analysis}
\label{subsec:cost}

Table~\ref{tab:cost} compares SSM-Attack against representative baselines along two hardware-independent axes: parameters required at inference, and the inference regime. Wall-clock timings are omitted because published runtimes come from heterogeneous hardware, ranging from a 40\,GB A100 to a four-GPU RTX~3090 node. For reference, on a single commodity NVIDIA~T4 our complete pipeline protects one image in $27$\,s at batch size $1$, which is the setting in which per-image runtimes are reported elsewhere, and $8.9$\,s per image when the surrogate forward passes are amortized over a batch of $16$. The allocation pass itself takes $0.16$\,s per image. The same T4 is used for training.

\begin{table}[t]
\centering
\caption{\textbf{Computational cost comparison.} ``Params'' counts, in millions, every
network evaluated at protection time, frozen ones included; surrogate FR models,
which all methods need equally, are excluded, and ``---'' marks methods with no
generator. Since the original papers do not report counts, they are summed from
the released architectures each method depends on: a $256^2$ DDPM U-Net for
DiffAM, CLIP~ViT-B/32 with a StyleGAN2 generator for CLIP2Protect, Stable
Diffusion for Adv-Diffusion and SDXL for Adv-CPG, while Adv-Makeup and AMT-GAN
are counted from the authors' code. Ours is measured directly: allocation
network $5.73$M, discriminator $2.77$M, truncated VGG-16 $7.64$M. $^{\S}$The
last two serve the quality terms only, so the attack-critical footprint is
$5.73$M. Wall-clock timings are omitted because published values come from
heterogeneous hardware.}
\label{tab:cost}
\setlength{\tabcolsep}{2.6pt}
\renewcommand{\arraystretch}{1.1}
\begin{tabular}{ll|r|l}
\toprule
Category & Method & Params & Inference regime \\
\midrule
\multirow{2}{*}{Noise}
& FGSM/PGD/MI-FGSM/TI-DIM & ---   & per-image optimization \\
& TIP-IM                        & ---   & per-image optimization \\
\midrule
\multirow{4}{*}{Makeup}
& Adv-Makeup                    & 0.58  & amortized forward pass \\
& AMT-GAN                       & 12.62 & forward pass, per-target training \\
& DiffAM                        & 113.67 & forward pass, per-target tuning \\
& CLIP2Protect                  & 181.65 & per-image latent search \\
\midrule
\multirow{3}{*}{\shortstack[l]{Sem.\\Inv.}}
& P3-Mask                       & ---    & amortized mask, per user \\
& DiffProtect                   & 168.49 & per-image latent optimization \\
& Adv-Diffusion                 & $\approx 1290$ & per-image latent diffusion \\
\midrule
Portrait
& Adv-CPG                       & $\approx 3500$ & per-image generation \\
\midrule
SSM
& \textbf{SSM-Attack (Ours)}    & \textbf{16.13}$^{\S}$ & amortized pass $+$ refinement \\
\bottomrule
\end{tabular}
\end{table}

The parameter gap is architecture-intrinsic: SSM-Attack requires 16.13M parameters in total, seven times smaller than DiffAM, eleven times smaller than CLIP2Protect and two orders of magnitude smaller than the diffusion- and portrait-based pipelines. It is not the smallest model compared: Adv-Makeup uses 0.58M parameters but perturbs only the eye-shadow region and attains 14.03\% ASR, and AMT-GAN's 12.62M generator is smaller than our full protection-time footprint, although it must be re-trained for every new target identity while ours is not.

The inference regime is equally consequential. Latent-search and diffusion methods solve a fresh optimization problem for every input, while AMT-GAN and DiffAM obtain fast forward-pass inference only by re-training the entire model for each new target identity, an amortized cost that per-image runtime conceals. SSM-Attack amortizes generation into a single trained network and adds a short projected refinement, so the refinement budget is a tunable efficiency lever rather than a fixed cost (Table~\ref{tab:ablation}k). The regime also determines what each count in Table~\ref{tab:cost} contains. One-forward-pass protection carries only the generator, whatever auxiliary networks training used: AMT-GAN trains with a learned perceptual metric, but its released inference path instantiates the generator alone. Per-image optimization carries the networks that optimization evaluates, which is why the latent-search and diffusion counts include their backbones and ours includes the discriminator and the feature extractor. The comparison is if anything conservative in their favor, since we cannot audit auxiliary terms in code we did not run.

The 16.13M figure ($16{,}129{,}574$ parameters) is the honest protection-time total and splits into three parts: the 5.73M allocation network, the 2.77M PatchGAN discriminator, and the 7.64M truncated VGG-16 feature extractor. The latter two are frozen but are evaluated during Stage~2, for the naturalness gradient of Eq.~\eqref{eq:nat_loss} and the perceptual term of Section~\ref{subsec:losses}, so by the accounting rule of Table~\ref{tab:cost} they belong in the count. Because both serve only the quality terms, a deployment that needs the attack alone can run the single-pass variant of Table~\ref{tab:ablation}(k) on the 5.73M allocation network, at $0.16$\,s per image.

\subsection{Ablation Study}
\label{subsec:ablation} Table~\ref{tab:ablation} isolates the contribution of the allocation mechanism.

\begin{table}[t]
\centering
\caption{\textbf{Ablation of the budget-allocation mechanism} on CelebA-HQ,
averaged over the four leave-one-out configurations. Every row except~(k) shares
the refinement pipeline and the cap $\epsilon_{\max}=0.10$; only the budget map
differs. ``Budget'' is the mean of $\boldsymbol{\epsilon}$ relative to uniform
allocation, i.e.\ allocated capacity, not the realised $\|\boldsymbol{\delta}\|$.
Rows marked $^{*}$ use the full 1{,}000-image test set; group~(A) a common
$150$-image subset and groups~(B) and~(E) a common $300$-image subset.
$^{\dagger}$Row~(k) uses the pre-refinement checkpoints, so it indicates what a
purely amortized variant delivers rather than being a controlled ablation of~(j).}
\label{tab:ablation}
\setlength{\tabcolsep}{3pt}
\renewcommand{\arraystretch}{1.08}
\begin{tabular}{cl|ccccc}
\toprule
& Configuration & ASR$\uparrow$ & SSIM$\uparrow$ & PSNR$\uparrow$ & FID$\downarrow$ & Budget \\
\midrule
\multicolumn{7}{l}{\emph{(A) Budget map}}\\
(a) & Uniform, $\epsilon_i \equiv \epsilon_{\max}$ & 88.17 & 0.8505 & 27.47 & 75.7 & 100\% \\
(b) & \textbf{SSM allocation} & 86.33 & \textbf{0.8943} & \textbf{29.09} & \textbf{64.5} & \textbf{37\%} \\
(c) & Elliptical face prior & 39.65 & 0.9455 & 33.53 & 29.2 & 17\% \\
(d) & SSM allocation $+$ generator init & 73.83 & 0.8752 & 28.13 & 109.1 & 37\% \\
\midrule
\multicolumn{7}{l}{\emph{(B) Allocation gain $\gamma$ (IR-152)}}\\
(e) & $\gamma = 1.0$ & 90.33 & 0.8910 & 29.03 & 60.3 & 40\% \\
(f) & $\gamma = 1.5$ & \textbf{97.67} & 0.8808 & 28.16 & 67.2 & 46\% \\
(g) & $\gamma = 2.5$ & 96.33 & 0.8795 & 28.09 & 66.6 & 47\% \\
(h) & $\gamma = 4.0$ & 96.67 & 0.8789 & 28.07 & 68.0 & 47\% \\
\midrule
\multicolumn{7}{l}{\emph{(C) Allocation gain, full test set}}\\
(i) & $\gamma = 1.0$\,$^{*}$ & 80.90 & \textbf{0.8957} & \textbf{29.11} & \textbf{48.8} & \textbf{39\%} \\
(j) & \textbf{Full pipeline, $\gamma{=}1.5$}$^{*}$ & \textbf{87.10} & 0.8878 & 28.42 & 53.6 & 42\% \\
\midrule
\multicolumn{7}{l}{\emph{(D) Refinement}}\\
(k) & No refinement, single generator pass$^{*\dagger}$ & 70.00 & 0.8670 & 27.00 & 150.6 & --- \\
\midrule
\multicolumn{7}{l}{\emph{(E) Allocation backbone (300-image subset)}}\\
(l) & \textbf{Mamba, raster scan (ours)} & \textbf{87.75} & \textbf{0.8851} & 28.37 & 63.7 & 38\% \\
(m) & Mamba, shuffled token order & 83.33 & 0.8798 & 28.37 & 63.5 & 39\% \\
\bottomrule
\end{tabular}
\end{table}

\noindent\textbf{Allocation shifts the attack--fidelity--budget trade-off.} Replacing the uniform map with the SSM-predicted allocation~(b) improves SSIM by $+0.044$, PSNR by $+1.62$\,dB and FID by $-11.2$ relative to~(a), while spending only $37\%$ of the budget a uniform allocation would use, at a cost of $1.84$ points of ASR. Because the saved budget can be reinvested through the allocation gain without violating the cap, the operating point~(j) reported in Table~\ref{tab:asr_celeba_ladn} attains $87.10\%$ ASR with every quality metric still ahead of uniform allocation. Under a fixed $\ell_\infty$ bound, \emph{where} the budget is spent therefore matters more than \emph{how much} is spent.

\noindent\textbf{The allocation is learned, not geometric.} Row~(c) replaces the learned map with a fixed elliptical face prior at comparable allocated capacity. ASR collapses to $39.65\%$, less than half the learned allocation, even though pixel-fidelity metrics improve: the prior simply spends less capacity, and spends it in a less useful place. Quantitatively the learned mask is close to binary: $61.9\%$ of pixels fall below $0.15$ of the cap and $35.9\%$ exceed $0.60$ (standard deviation $0.466$), whereas a mask trained with only a mean-value regularizer degenerates to a spatially flat map (standard deviation $0.059$) and behaves indistinguishably from~(a). The coverage and selectivity terms of Section~\ref{subsec:losses} prevent this degeneration.

\noindent\textbf{Where the budget goes.} Over 200 CelebA-HQ images, a central face ellipse covering $38.7\%$ of the frame receives $74.4\%$ of the allocated capacity ($1.93\times$ its area share) and the periocular--nose--mouth region, covering $12.1\%$, receives $25.9\%$ ($2.14\times$). The remaining quarter follows the hair boundary and jaw silhouette rather than spreading over the background, consistent with FR embeddings encoding outline cues.

\begin{figure}[t]
\centering
\includegraphics[width=\textwidth]{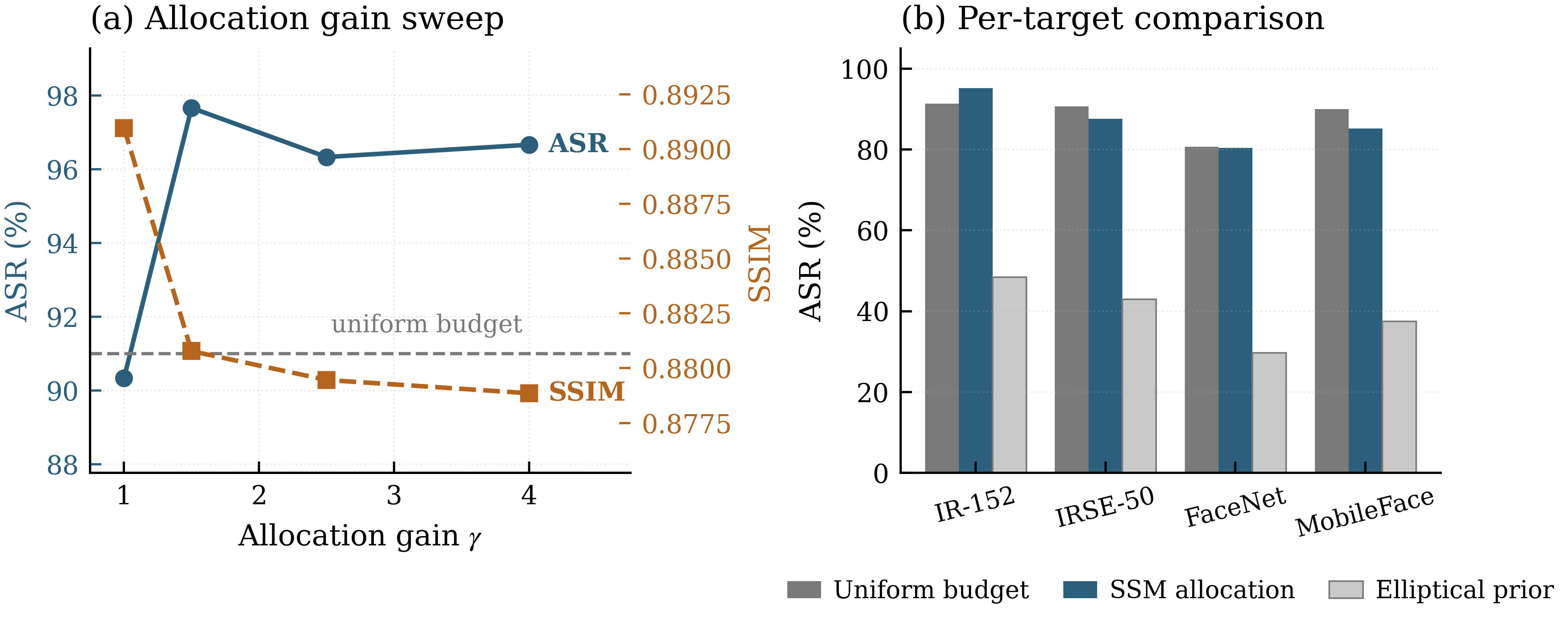}
\caption{\textbf{Budget allocation under a fixed $\ell_\infty$ cap.}
\textbf{(a)} Allocation gain sweep on the IR-152 configuration; the dashed line
marks uniform allocation at the same cap. We fix $\gamma = 1.5$, the knee of the
curve, for all reported results.
\textbf{(b)} Per-target comparison at $\gamma{=}1.5$: learned allocation beats
uniform on the two hardest targets while using $42\%$ of its capacity, whereas
an elliptical prior at comparable capacity recovers less than half.}
\label{fig:eps_tradeoff}
\end{figure}

\noindent\textbf{The network should allocate, not synthesize.} Row~(d) additionally initialises $\boldsymbol{\delta}$ from a generator-synthesized perturbation. ASR drops by $12.50$ points and FID increases by $44.6$. Spectral analysis of the synthesized output reveals a periodic component at the $16$-pixel period of the token grid; replacing transposed convolutions with bilinear upsampling reduces this peak from $7.96\times$ to $0.66\times$ the neighbouring bands, yet the residual image-independent structure still degrades the distributional metric, whereas gradient-determined perturbation values do not.

\noindent\textbf{Allocation gain and refinement budget.} Group~(B) sweeps $\gamma$ on the IR-152 configuration (Fig.~\ref{fig:eps_tradeoff}a): ASR rises sharply from $\gamma{=}1.0$ to $\gamma{=}1.5$ and then saturates, because the near-binary mask recruits no additional pixels once the cap binds; we fix $\gamma = 1.5$ throughout. Group~(C) repeats the two endpoints over all four configurations on the full test set: $\gamma{=}1.0$ trades $6.20$ points of ASR for the best quality of any refined configuration (SSIM $0.8957$, PSNR $29.11$, FID $48.8$), so the gain is itself an operating-point control. Group~(D) drops the refinement and protects with a single generator pass in $0.16$\,s per image ($\approx 170\times$ faster at batch size $1$). It keeps $70.00\%$ ASR, four fifths of the full pipeline, but FID rises to $150.6$: without gradient-determined values the perturbation carries the image-independent structure diagnosed in row~(d), undiluted. An amortized variant is therefore viable where latency dominates, while refinement is what makes the output distributionally acceptable.

\noindent\textbf{The token ordering is load-bearing.} Group~(E) applies a fixed random permutation to the token sequence at protection time with every weight unchanged; were the ordered scan uninformative the result would not move. ASR falls by $4.42$ points over the pooled $1{,}200$ protections ($300$ images per target; $p = 0.002$) at unchanged PSNR and FID, so nothing is traded for quality. The loss is concentrated rather than uniform: the other three targets lose $1.78$ points together, which the data do not separate from zero ($p = 0.24$), whereas FaceNet loses $12.33$ ($p = 0.0004$, surviving Bonferroni correction for four comparisons). FaceNet is also the configuration in which the target is furthest from its surrogates, being the only InceptionResNet backbone with a triplet objective while its three surrogates are margin- or softmax-trained residual networks; ordering therefore appears to matter where transfer is hardest, though a single target does not establish that as a general rule. Two caveats bound the claim. Because the permutation also disrupts the causal convolution inside the Mamba block, the evidence supports the ordered token sequence as a whole rather than the state-space recurrence in isolation. And because both arms are evaluated on the same images, the observations are paired; we report the unpaired two-proportion test, which is the more conservative of the two.

We use this permutation rather than a comparison against convolutional or attention-based allocators because it isolates the property in question. Substituting a different backbone changes architecture, parameter count, optimization difficulty and tuning effort at the same time, so any difference it produces cannot be attributed to sequence modeling alone, and a control trained with less care than the proposed model would bias the comparison in our own favor. The permutation changes exactly one thing: the order in which the same weights visit the same tokens. Weights, capacity, training data, refinement schedule and budget cap are identical across the two arms, and the permutation is applied only at protection time. What the experiment therefore establishes is that the ordered token sequence carries information the allocation uses, not that a selective SSM allocator outperforms a convolutional or attention-based one; that comparison remains open.

\section{Discussion and Limitations}
\label{sec:discussion_limitations} SSM-Attack closes the transferability gap between noise-based and unrestricted methods through allocation rather than capacity: a 5.73M-parameter allocation network applying strictly $\ell_\infty$-bounded noise attains the highest average ASR on all three evaluation sets while spending under half the budget of a uniform map at the same cap.

The mechanism is interpretable. Modern FR systems encode identity as a holistic spatial direction~\citep{deng2019arcface}, and a selective state update carries context along the token sequence in one pass, so a decision made late in the scan can draw on far-apart regions of the face; Table~\ref{tab:ablation}(E) is the direct evidence. The \textit{untargeted} objective also spends the whole budget on identity suppression, whereas targeted attacks often dodge far worse than they impersonate~\citep{zhou2024rethinking}.

\noindent\textbf{Cross-dataset generalization.} The CelebA-HQ and LADN results come from networks trained on FFHQ$+$MT, disjoint from both evaluation sets, one per held-out target (Table~\ref{tab:protocol}). Networks trained instead on CelebA-HQ$+$MT attain $85.99\%$ on LADN, $0.91$ points below, and $81.15\%$ on the unseen FFHQ corpus (Table~\ref{tab:ffhq}). The allocation therefore transfers across datasets; establishing dataset-independence would need more corpora than the three evaluated here.

\noindent\textbf{Perturbation budget.} The per-model budgets are set by the surrogate ensemble, not by knowledge of the held-out target; a universal $\epsilon{=}0.10$ can be used in deployment.

\noindent\textbf{LADN protocol.} Unlike CelebA-HQ, LADN has no shared evaluation subset, and for the same method DiffAM~\citep{sun2024diffam} and the supplementary material of Adv-CPG~\citep{wang2025advcpg} report LADN scores differing by up to $19$ points on individual FR models. We therefore state our protocol ($332$ non-makeup images, hyperparameters identical to CelebA-HQ) and, where sources disagree, quote the higher value, so competing methods appear at their strongest published numbers.

Several limitations remain. Transferability depends on surrogate diversity and may degrade against fundamentally new FR paradigms, and our evaluation is restricted to static images and the four FR models of the standard benchmark; commercial recognition services, which several compared methods report, are not evaluated. Projected refinement adds cost over the single allocation pass; its iteration count is a deployment-time lever (Table~\ref{tab:ablation}k). FID remains above manifold-editing methods: allocation reduces it by $11.2$ points relative to a uniform map but does not close a gap that is intrinsic to additive perturbation. The scan is unidirectional and follows a raster order, and the significant ordering effect of Section~\ref{subsec:ablation} is confined to one held-out target, so neither the choice of traversal nor the generality of the ordering effect is settled here. Finally, naturalness is imposed by loss terms rather than structurally, as in manifold-editing methods; a structural formulation would remove the frozen perceptual network from the protection-time footprint. Future work could explore bidirectional spatial modeling~\citep{zhu2024visionmamba}, defense-aware training against purification, surrogate diversification~\citep{zhou2025dpa}, and allocation for video, where the map must stay temporally consistent.

\section{Conclusion}
\label{sec:conclusion} We introduced SSM-Attack, the first adversarial face privacy framework based on Selective State Space Models, which uses the linear-time scan of Mamba to decide where a strict $\ell_\infty$ budget is spent across the face. With an allocation network of only 5.73M parameters, it achieves the highest average ASR on CelebA-HQ (87.10\%), LADN (86.90\%) and FFHQ (81.15\%), which suggests that learned budget allocation is a general design principle for bounded adversarial attacks.

\section*{CRediT authorship contribution statement}
\textbf{Tunahan Parlay{\i}c{\i}:} Conceptualization, Methodology, Software, Validation,
Investigation, Data curation, Visualization, Writing -- original draft.
\textbf{Erchan Aptoula:} Supervision,
Writing -- review \& editing.
\textbf{Y\"{u}cel Sayg{\i}n:} Supervision, Funding acquisition, Writing -- review \& editing.
\section*{Declaration of competing interest}
The authors declare that they have no known competing financial interests or personal relationships that could have appeared to influence the work reported in this paper.
\section*{Data availability}
Source code and pretrained models will be released at \url{https://github.com/7unahan/PrivacySSM} upon acceptance. All datasets used are publicly available.
\section*{Funding}
This work was supported by the Scientific and Technological Research Council of T\"{u}rkiye (T\"{U}B\.{I}TAK) under the 2244 Industrial Ph.D. Program. The funder had no role in study design, analysis, the decision to publish, or preparation of the manuscript.
\section*{Declaration of generative AI and AI-assisted technologies in the manuscript preparation process}
During the preparation of this work the authors used Claude (Anthropic) and Grammarly in order to improve the readability and language of the manuscript, and Claude to assist with the draw.io layout scripts for the graphical abstract, which the authors designed and edited. After using these tools, the authors reviewed and edited the content as needed and take full responsibility for the content of the published article.

\bibliographystyle{elsarticle-num}
\bibliography{cas-refs}
\end{document}